\documentclass{article}

\usepackage{arxiv}

\usepackage[utf8]{inputenc} 
\usepackage[T1]{fontenc}    
\usepackage{hyperref}       
\usepackage{url}            
\usepackage{booktabs}       
\usepackage{amsfonts}       
\usepackage{nicefrac}       
\usepackage{microtype}      
\usepackage{lipsum}
\usepackage{graphicx}
\usepackage{amsmath}
\usepackage{subfig}

\title{Reproduction of Lateral Inhibition-Inspired Convolutional Neural Network for Visual Attention and Saliency Detection}

\author{
  Filip Marcinek \\
  University of Wrocław \\
  \texttt{282905@uwr.edu.pl} \\
}

\begin{document}
\maketitle

\begin{abstract}
In recent years, neural networks have continued to flourish, achieving high efficiency in detecting relevant objects in photos or simply recognizing (classifying) these objects - mainly using CNN networks. Current solutions, however, are far from ideal, because it often turns out that network can be effectively confused with even natural images examples \cite{adversarial}. I suspect that the classification of an object is strongly influenced by the background pixels on which the object is located.  In my work, I analyze the above problem using for this purpose saliency maps created by the LICNN network. They are designed to suppress the neurons surrounding the examined object and, consequently, reduce the contribution of background pixels to the classifier predictions. My experiments on the natural and adversarial images datasets show that, indeed, there is a visible correlation between the background and the wrong-classified foreground object. This behavior of the network is not supported by human experience, because, for example, we do not confuse the yellow school bus with the snow plow just because it is on the snowy background.
\end{abstract}

\section{Introduction}
Research on how strongly the background pixels affect the prediction of neural networks and how this problem can be remedied is very important in the context of the contemporary development of neural networks, the widespread acceptance of such solutions in the field of the production of everyday objects as well as autonomous cars in the future. Especially in this latter aspect, it is important for us that the neural network identifying objects around the car can recognize difficult cases such as pedestrians appearing on the road (although this is not their natural background). If the pavement as a background had a big impact on the classification of the "people" category, there was a great danger of categorizing objects on the pavement as people or, which is worse, people on the road as something else. The above motivation seems strong enough to take a closer look at the issue of background significance in neural network predictions - and if this impact proves to be too huge, to examine whether it can be reduced in any way.

In my work, I will first provide a general outline of the CNN network, which has so far been the most successful in the field of visual challenges and is also widely used today. Next, I will introduce the concept of saliency map and present how it can be useful in the topic of neural networks. In the next chapter I will present the LICNN network that will allow us to create good-quality saliency maps, I will discuss how I implemented the above network and show how it looks as compared to other similar methods. In the subsequent chapter I present my experiments and their results. The last chapter contains conclusions resulting from the conducted experiments. Source code could be found here\footnote{\url{https://github.com/fmarcinek/LICNN}}.

\subsection{General outline of CNN}

Convolutional Neural Network (CNN) is a well-known deep learning architecture, which is oriented towards processing data with known grid topology e.g. graphic data that can be represented as a two-dimensional pixel grid. It is successfully used in many computer vision tasks including image classification, visual saliency detection or pose estimation \cite{advances}.

\begin{figure}[h!]
\centering
\includegraphics[scale=0.27]{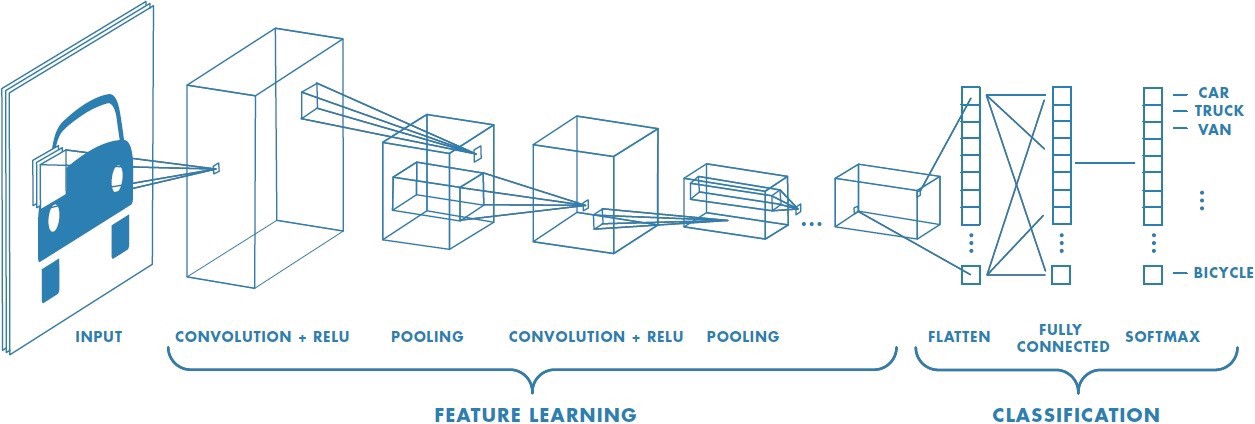}
\caption{CNN structure scheme \cite{bib:cnn_scheme}.}
\label{fig:cnn_structure}
\end{figure}

The structure of this network (Fig. \ref{fig:cnn_structure}) was based on discoveries of the process of visual perception in mammals, in which it was noted that the neurons from the first segments of the visual cortex react most strongly to very specific patterns, e.g. lines with precise orientation. Then these patterns activated in small areas are processed in more complex cells, which remain unchanged into small shifts of the feature. The above neurobiological concepts have been transformed into the corresponding (at least partially): convolutional layer and pooling layer, which I will discuss below with all other CNN building blocks.

\begin{figure}[h!]
\centering
\includegraphics[scale=0.27]{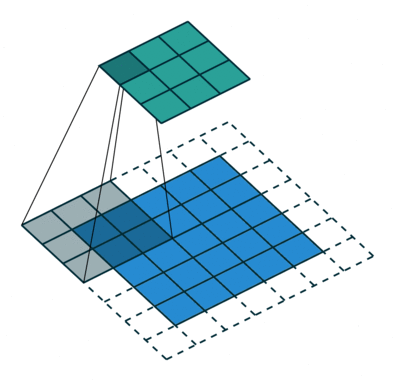}
\caption{Convolutional layer operation \cite{bib:convolve}.}
\label{fig:conv}
\end{figure}

\textbf{Convolutional layer} \\
The convolutional layer aims to learn feature representations of the inputs and is composed of several convolution kernels which are used to compute different outputs (named feature maps), for example each channel from three-channeled image contains different features so we could use three different kernels to activate different patterns in each channel. Specifically, each neuron of a feature map (output of convolutional layer) is connected to a region of neighbouring neurons in the convolutional layer input what was depicted on Fig. \ref{fig:conv}.

Such a neighbourhood is referred to as the neuron’s receptive field in the input layer. The new feature map can be obtained by at first convolving the input with a learned kernel and then applying on the result an element-wise nonlinear activation function. Feature maps may be gained by using several different kernels (for example when there is one kernel for each input channel). Mathematically, the feature value $y_{ij}$ in the \textit{k}-th feature map, is calculated by:

\begin{equation}
y_{ij, k} = sum\_all(\boldsymbol{w}_k \odot \boldsymbol{x}_{ij}) + b_k
\end{equation}
where \\ \hspace*{1cm} $\boldsymbol{w}_k$ and $b_k$ mean \textit{k}-th kernel containing weights and its bias term,  \\ \hspace*{1cm} $\boldsymbol{x}_{ij}$ is the input patch centered at location (i, j), \\ \hspace*{1cm} $\odot$ means Hadamard product of two matrices, \\ \hspace*{1cm} \textit{sum\_all()} means summing all values from matrix.

It is worth to notice that the kernel generating feature map have weights shared for each input. Weight sharing reduces the model complexity, hence cause the model easier to train (gives some computational benefits, e.g. the weights can be stored in cache).

\textbf{Nonlinearity layer} \\
Convolutional layers are followed by nonlinearities using the activation function, which is applied to all values in the feature maps. It introduces nonlinearity to CNN, which allows deep neural networks to learn nonlinear dependencies in the data. Typical activation functions are \textit{ReLu}, \textit{sigmoid} or \textit{tanh}.

\begin{figure}[h!]
\centering
\includegraphics[scale=0.4]{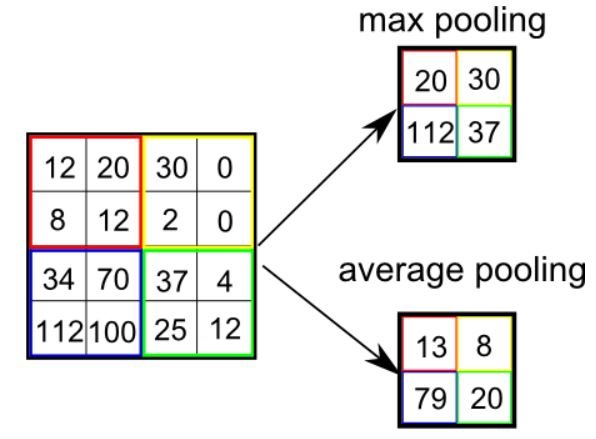}
\caption{Max-pooling and average-pooling examples \cite{bib:pool}.}
\label{fig:pooling}
\end{figure}

\textbf{Pooling layer} \\
The pooling layer does no learning itself (does not have any weights). It endeavors to gain shift-invariance by reducing the feature maps resolution, which is performed by extracting simple activation statistics from the local scope of the neuron map. The most commonly used statistics are maximum and arithmetic average functions (Fig. \ref{fig:pooling}). The pooling layer simply take some $k$ x $k $ patch and output a single value, which is a certain linear operation. For instance, if their input layer is a $N$ x $N$ layer, they will then output a $\frac{N}{k}$ x $\frac{N}{k}$ layer, as each $k$ x $k$ block is reduced to just a single value.

\textbf{Fully-connected layer} \\
After convolutional layers (sometimes interleaved with pooling layer) occur fully-connected layers which perform high-level reasoning. Inside these layers there are connections between every input-output neurons pair. After each fully-connected layer there is also a nonlinearity layer, which activate only neurons with enough strong signal. 

\textbf{Loss layer} \\
The last layer of CNNs is an loss layer. Because of training CNN is a global optimization problem, we can obtain the best fitting parameters by searching the global minimum of the loss function. For classification tasks, the \textit{softmax} function is commonly used. Let $\boldsymbol{\theta}$ denote all the parameters of CNN, e.g. kernels and bias terms. When we have N desired input-output relations, $\boldsymbol{y}^{(i)}$ is target label corresponding \textit{i}-th input and $\boldsymbol{o}^{(i)}$ is the output of CNN, the loss of CNN can be calculated as following average:
\begin{equation}
\mathcal{L} = \frac{1}{N} \sum_{n=1}^N \ell(\boldsymbol{\theta}; \boldsymbol{y}^{(n)}, \boldsymbol{o}^{(n)})
\end{equation}
The basic CNN components I have presented enable capturing two-dimensional dependencies, which makes this network an effective tool in image processing.

\subsection{Saliency maps}
Saliency map is a popular tool to highlight features in an input data considered important for the concrete prediction of a learned model \cite{sanity}.  Saliency map is often also called attention map. The goal of saliency maps is to visualize which portions of input data are most crucial to the model and which are less. They are widely used to debug predictive models as well as to explaining specific model decisions \cite{bib:saliency2}.

\begin{figure}[h!]
\centering
\includegraphics[scale=0.36]{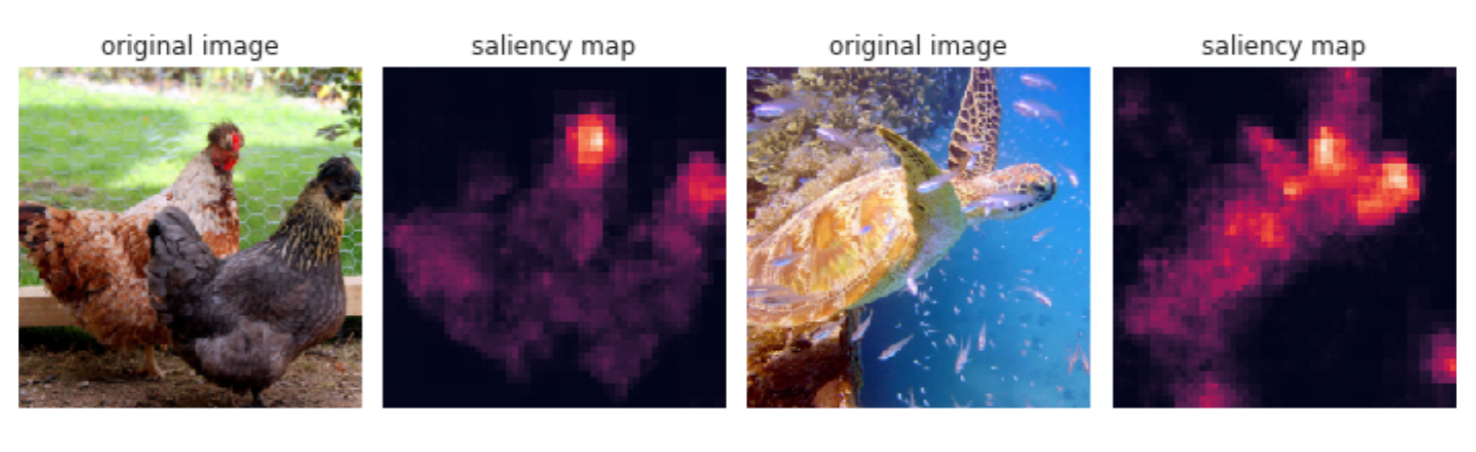}
\caption{Saliency maps examples created by my LICNN implementation.}
\label{fig:saliency_example}
\end{figure}

\section{LICNN}
\subsection{LICNN overview}
\begin{figure}[h!]
\centering
\includegraphics[scale=0.24]{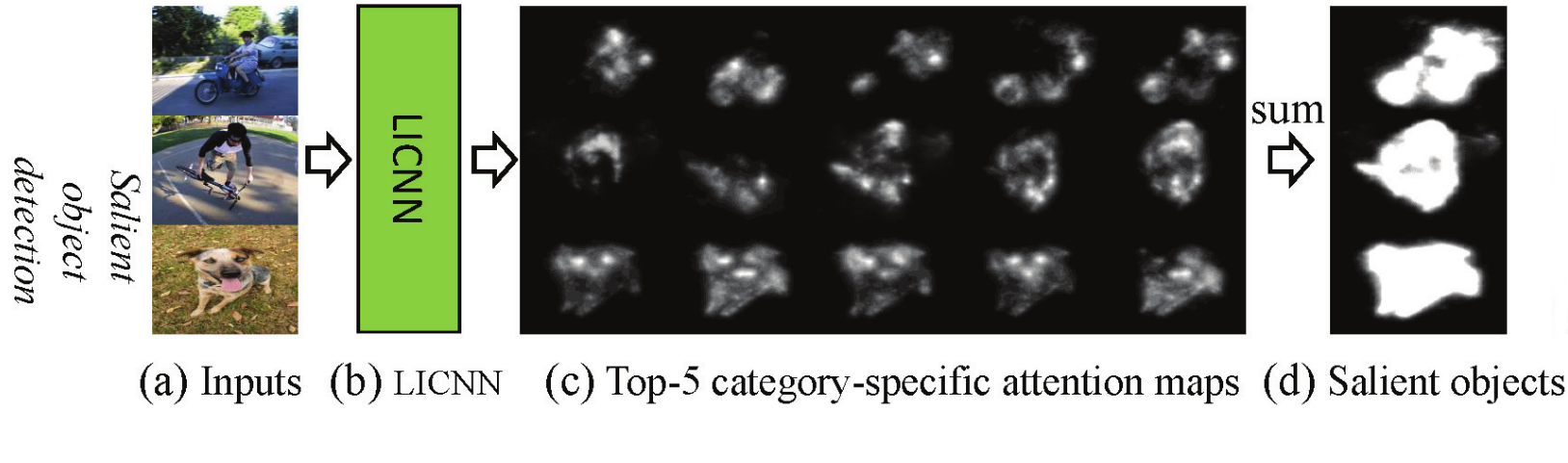}
\caption{Scheme of creating saliency maps from the paper \cite{cao}.}
\label{fig:saliency_creation}
\end{figure}

Lateral Inhibition CNN (LICNN) is an attempt of modeling visual attention, have been proposed by Chunshui Cao \textit{et al.} in their paper on \textit{AAAI-18 Conference}. As the authors ensure LICNN could classify image, produce attention map for the specific category and also create saliency map from the given image. We would like to gain a saliency map for the input image, what in the paper is simply reduced to obtaining top-5 category-specific attention maps (maps which highlight concrete object category) and then summing them together and normalizing (Fig. \ref{fig:saliency_creation}). 

\begin{figure}[h!]
\centering
\includegraphics[scale=0.5]{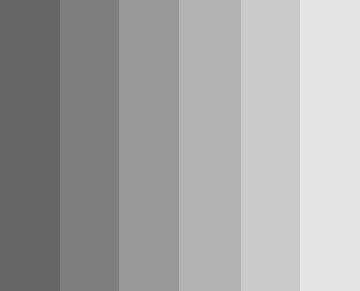}
\caption{Mach bands as an example of lateral inhibition \cite{bib:mach_bands}.}
\label{fig:mach}
\end{figure}

\textbf{Lateral inhibition concept} \\
LICNN authors are inspired by neurobiological concept of lateral inhibition which occurs e.g. in the human brain and is "a type of process in which active neurons suppress the activity of neighboring neurons through inhibitory connections" \cite{cao}. This behaviour should therefore reinforce neurons with higher signal, blocking ones weaker at the same time. Good example of this neurons behaviour can be seen in the presented picture of Mach bands (Fig. \ref{fig:mach}). It seems that along the edge between neighboring grey bands, lateral inhibition makes an illusion that the darker area near the border seems even darker and the lighter seems even lighter. In fact, the bands have a uniform color.

\textbf{Top-down feedback information using} \\
The authors rely on the achievements of earlier scientists who has already adopted lateral suppression in their networks and showed that "lateral suppression by neighboring neurons in the same layer makes the network more stable and efficient" \cite{cao}. However, the previous attempts used lateral inhibition mechanism during the bottom-up procedure, whereas the authors of LICNN inspire by neurobiology again and apply the above mechanism to top-down gradient feedback signals because "top-down attention plays an important role in human visual system" \cite{cao}.

Their idea is to use feedback gradient signal to estimate how much pattern learned by the neuron, contributes to the given category. This concept is based on the observation that "specific neurons will be activated if there are positively correlated patterns lying in their receptive field. These activated patterns give rise to distinct scores of different classes" \cite{cao}. As it was calculated in the paper, quantitatively estimation how much a neuron belongs to the specific category can be simply computed by back-propagation from the given category. Similar solutions using back-propagation to create category-specific attention map have already been proposed \cite{bib:saliency}.

It is important to note that the saliency maps are extracted using a CNN trained on the image labels, so no additional information is required (such as object bounding boxes or segmentation masks). The computation of the image-specific saliency map for a single class is extremely quick, since it only requires a single back-propagation pass.

\textbf{Lateral inhibition model} \\
In this moment we can obtain a vague and not so precise attention map for a concrete category (using category-specific gradient), so authors propose to apply lateral inhibition model to obtained map to "suppress noise and increase the contrasts between target objects and background" \cite{cao}. These results will be gained by applying an inverted "Mexican hat" function, which is used in neurobiology as a computational model, to 
this will be discussed in Section (2.2).

\begin{figure}[h!]
\centering
\includegraphics[scale=0.3]{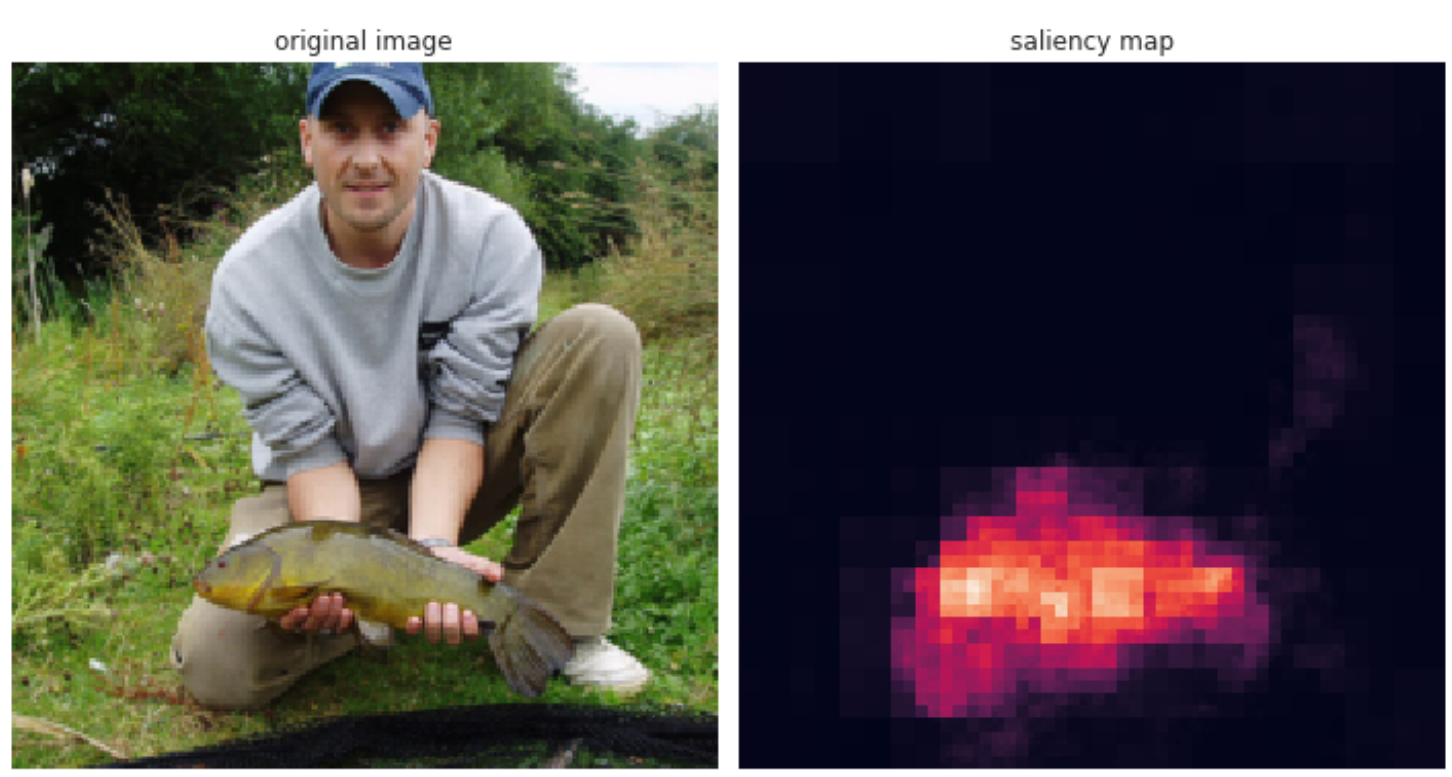}
\caption{Example of lack salient object on saliency map.}
\label{fig:no_people_img}
\end{figure}

\textbf{Observations about saliency maps produced by LICNN} \\
Authors claim that interesting capacity of LICNN is that it can "effectively locate salient objects even when the input image does not contain predefined objects in the CNN classifier" \cite{cao}. Unfortunately, the above assumption has not been fully confirmed, because my experiments show that e.g. people are not recognized by LICNN as salient objects despite the fact that they are definitely located in the foreground (Fig. \ref{fig:no_people_img}). The reason may be a fact that ImageNet dataset does not contain "man" nor "people" class, so networks learned on ImageNet just ignore people on the image -- this leads to the conclusion that the above problem occurs only in this type of network and is expected. I think it is true that CNN could have learned many visual local patterns which are shared by different objects (even those category did not participate in network training) and thus can highlight these shared patterns in the output saliency map.

\subsection{LICNN implementation details}
\begin{figure}[h!]
\centering
\includegraphics[scale=0.27]{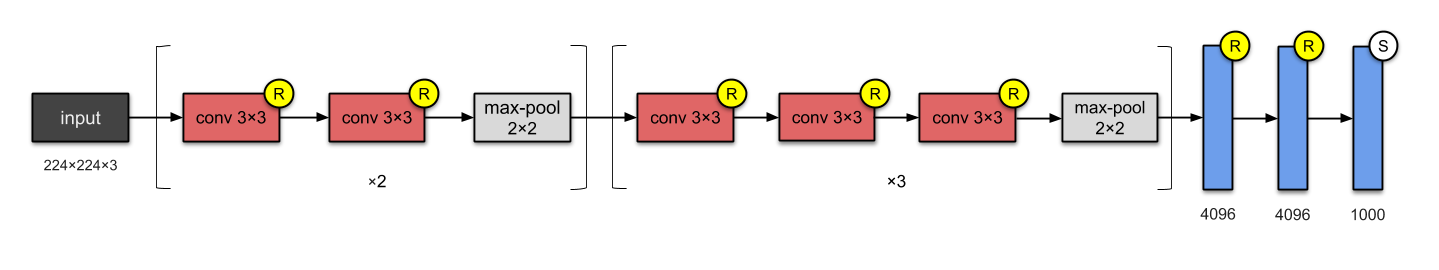}
\caption{VGG-16 structure \cite{bib:vgg_structure}.}
\label{fig:vgg16}
\end{figure}

\textbf{VGG-16 details} \\
This network uses VggNet architecture (VGG-16) pre-trained on ImageNet2012 dataset \cite{imagenet}, adopted from Caffe Model Zoo. In my implementation I used the same architecture but written and pretrained by PyTorch in its library. Network is used in evaluation mode. The architecture of VGG-16 is as follows (Fig. \ref{fig:vgg16}): a fixed-size 224x224 RGB image is passed through 13 convolutional layers, where there are used filters with a very small receptive field 3x3. After each convolutional layer there is ReLu layer which remains only neurons with positive values. Spatial pooling is performed by 5 max-pooling layers, which follow some of the convolutional layers as shown in the picture (Fig. \ref{fig:vgg16}). Max-pooling is performed over a 2x2 pixel window with stride 2. At the end there are 3 fully-connected layers: first two have 4096 channels and end up with ReLu; the third performs 1000-way ImageNet classes classification and thus contains 1000 channels (one for each class) and ends up with softmax layer, which calculates 1000 outputs probabilities.

The input to LICNN (as in VGG-16) is a fixed-size 224x224 RGB image. The only preprocessing is subtracting from each pixel the mean RGB value (computed on the training ImageNet set) and dividing each pixel by standard deviation (obtained in the same way). If an image is larger than 224x224, then first I change its size in manner that the shorter edge of the image is 256, and then I crop the image to 224x224.

\textbf{Lateral inhibition model implementation} \\
For matrix of dimensions $(W,H,C)$, where $W$, $H$, $C$ denote width, height and channels respectively, in the following discussion:
\begin{itemize}
\item  Max-C map -- means matrix of dimensions $(W, H)$ obtained by simple \textit{max} operation on all channel values in each $(w,h)$ location;
\item Sum-C map -- means matrix of dimensions $(W, H)$ obtained by simple \textit{sum} operation on all channel values in each $(w,h)$ location.
\end{itemize}

The image is normally passed through the network except that I save returned activation layer tensors from each ReLu output for further lateral inhibition computations.  When VGG returns class probabilities, after softmax execution[see Apendix], I perform back-propagation for top-1 (or for each from top-5 separately if I want to obtain saliency map) class nodes using automatic differentiation package from PyTorch. In this manner I gain contribution weight to the given class for every neuron in the network. Then I can get to the gradient values (category-specific contribution weights) through saved activation layer tensors. For each ReLu, I calculate its Max-C map, then I compute inhibition strength for each location in the following manner.

Let $x_{ij}$ denotes a point in the Max-C map on $(i,j)$ location and then lateral inhibition value will be denoted as $x_{ij}^{LI}$. It is calculated in $k$ x $k$ square patch called Lateral Inhibition Zone (LIZ), $k$ is a length of LIZ. LIZ consists of $k$ x $k$ nearest neighbours of $x_{ij}$ in Max-C map, where $x_{ij}$ is the centre of LIZ. $x_{ij}^{LI}$ result can be written formally as:
\begin{equation}
x^{LI}_{ij} = a \cdot \underbrace{e^{-\overline{x_{ij}}}}_{average} + b \cdot \underbrace{\sum_{uv} (d_{uv}e^{-d_{uv}}\delta(x_{uv} - x_{ij}))}_{differentiation}
\end{equation}
where \\ \hspace*{1cm} $\overline{x_{ij}}$ is a mean of all values in LIZ,
\\ \hspace*{1cm} $x_{uv}$ is a neighbour of $x_{ij}$ in LIZ,
\\ \hspace*{1cm} $d_{uv}$ is the Euclidean distance between $x_{ij}$ and $x_{uv}$, divided by $k$,
\\ \hspace*{1cm} $\delta(x) = max(0,x)$,
\\ \hspace*{1cm} $a$ and $b$ are the balance coefficients.

As they write in the paper: "The average term protects the neurons within a high response zone. And the differential term sharpens the objects’ boundaries and increases the contrast between objects and background in the protected zone created by the average term." \cite{cao}

For parameters in Equation (3) I used values given in the LICNN paper: $a = 0.1$, $b = 0.9$ and $k = 7$ (selected intuitively by the authors). It should be calculated lateral inhibition matrix of the same dimensions as Max-C map, during computations should be used zero-padding equal to $\lfloor k / 2\rfloor$.

When we obtain lateral inhibition matrix, we should normalize it with L2 norm and then we get final suppression mask by perform gating operation:

\begin{equation}
x_{ij} = \begin{cases}
x_{ij} &\text{if \(  x_{ij} - x_{ij}^{LI} > 0 \)}  \\
0 &\text{else}
\end{cases}
\end{equation}
Now, when we get suppression masks from all ReLu layers, we perform feed-forward of the VGG again, during which after each ReLu layer of dimensions $(W,H,C)$ we erase (assign $0$ value) all channels in all locations $(w,h)$ where suppression mask for this ReLu layer is $0$.

\textbf{Algorithm scheme} \\
Scheme of the algorithm for obtaining attention map:
\begin{enumerate}
\item Given an image and pretrained CNN classifier.
\item Perform feed-forward and gain predicted category.
\item Carry out gradient back-propagation of the predicted category to estimate contribution to the given category for all neurons.
\item  In each gained by back-propagation ReLU layer, compute the Max-C map and apply the lateral inhibition model on the obtained Max-C map (save obtained suppression mask to the future feed-forward).
\item  Perform feed-forward again, during which in each ReLu output layer erase through all channels locations which are 0 in suppression mask for this ReLu layer (erase means 'assign 0 value to it').
\item Calculate Sum-C map and normalize with L2 norm for each ReLu output layer (activation layer) obtained in this feed-forward.
\item Resize all Sum-C maps to the input image size, sum all together and normalize with L2 norm.
\end{enumerate}

To gain a saliency map, one should perform the above algorithm separately for each of top-5 predicted categories, then sum all these category-specific attention maps together and normalize with L2 norm.

\subsection{Comparison between LICNN results from the paper and mine}

The authors did not provide any repository with images used in paper, which is why I obtained the images by cutting them out of from the printscreen of the paper. It can be seen that the Dalmatians are located extremely well (in this case the network was asked to create category-specific attention map for the given Dalmatian class)\footnote{Difference in the resolutions of the presented images is due to the fact that my attention maps have the same size as the input to the network (224x224), while these from paper are imposed on the original images. Black frame indicates the input area.}. 

\begin{figure}[h!]
\centering
\includegraphics[scale=0.5]{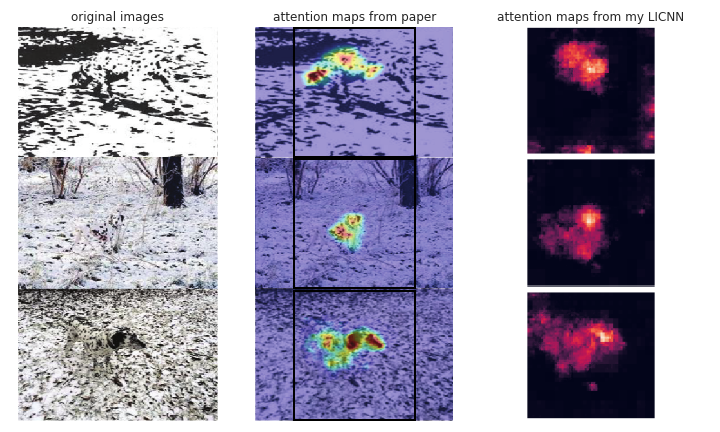}
\caption{Comparison between paper's attention maps and mine. Black frame indicates the attention map area.}
\end{figure}

\newpage

\section{Comparison of LICNN with other saliency map estimation methods}
\subsection{Sanity check of LICNN}
\textbf{Model parameter randomization test} \\
I have implemented and conducted the model parameter randomization test proposed by scientists from Google Brain and Berkeley University, which should find whether saliency map creation method could be used to explaining the relationship between inputs and outputs that the model learned and debugging the model \cite{sanity}. The model parameter randomization test saliency method by comparison between saliency map obtained by a trained model and a saliency map gained by randomly initialized network of the same architecture. If the saliency method relies on the learned parameters of the neural net, we can expect that the saliency maps obtained from these extreme different cases will differ significantly. However, in the case when we get similar maps, we can deduce that given saliency map is insensitive to the model parameters. In particular, this saliency map cannot be useful for tasks such as model debugging which unavoidably rely on the model parameters. 

\begin{figure}[h!]
\centering
\includegraphics[scale=0.45]{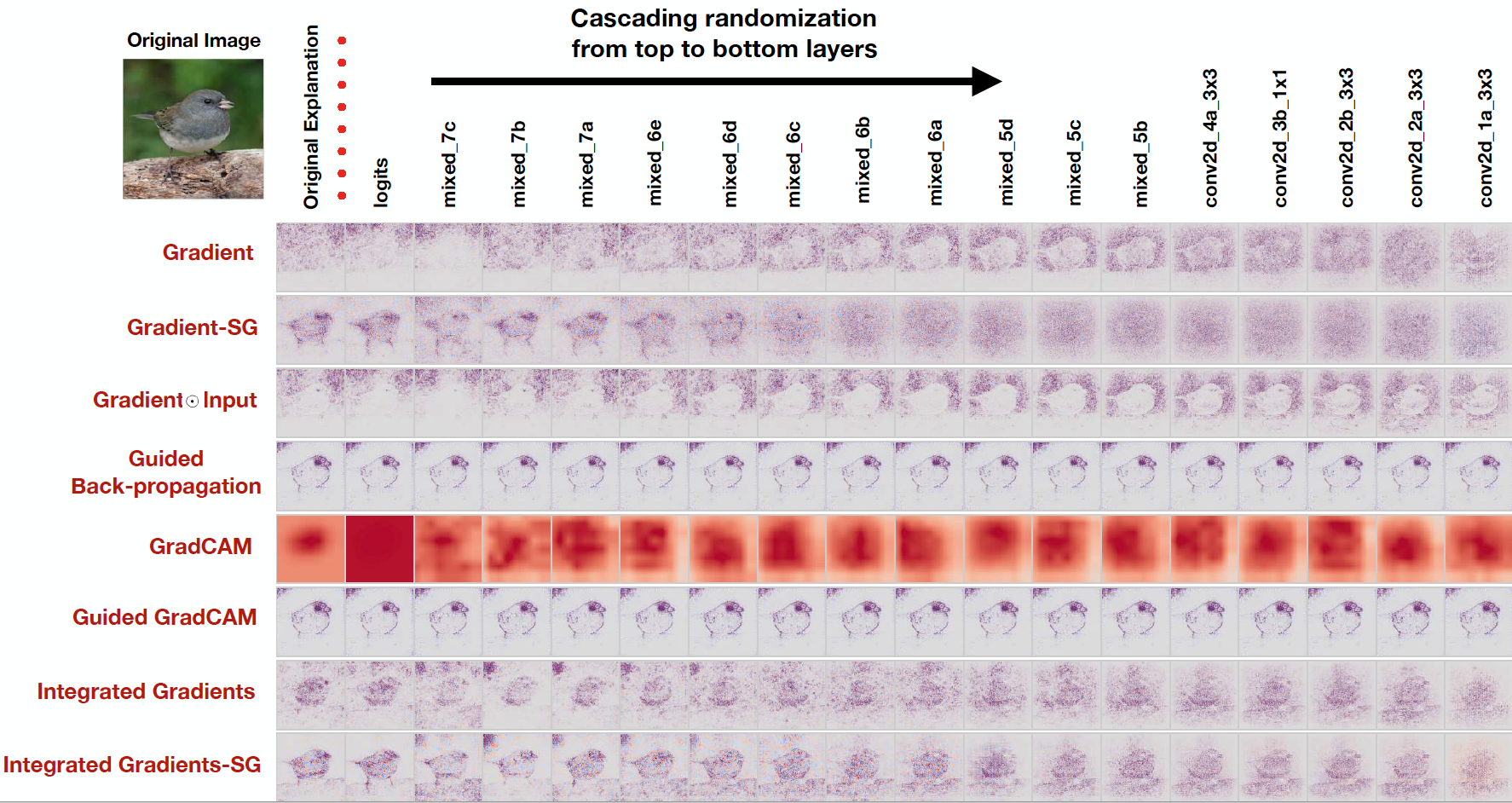}
\caption{Saliency maps from original experiment on several saliency methods. It shows subsequent maps from cascading randomization test (from left to right) \cite{sanity}.}
\label{fig:cascading}
\end{figure}

\begin{figure}[h!]
\centering
\includegraphics[scale=0.19]{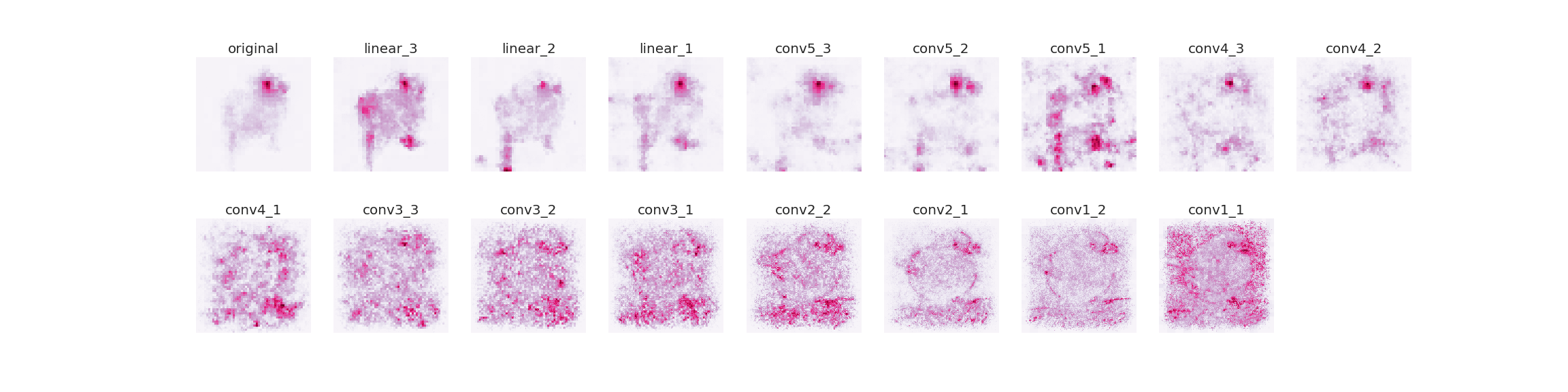}
\caption{Cascading randomization resulting saliency maps of the Junco bird image.}
\label{fig:my_casc}
\end{figure}

\begin{figure}[h!]
\centering
\includegraphics[scale=0.4]{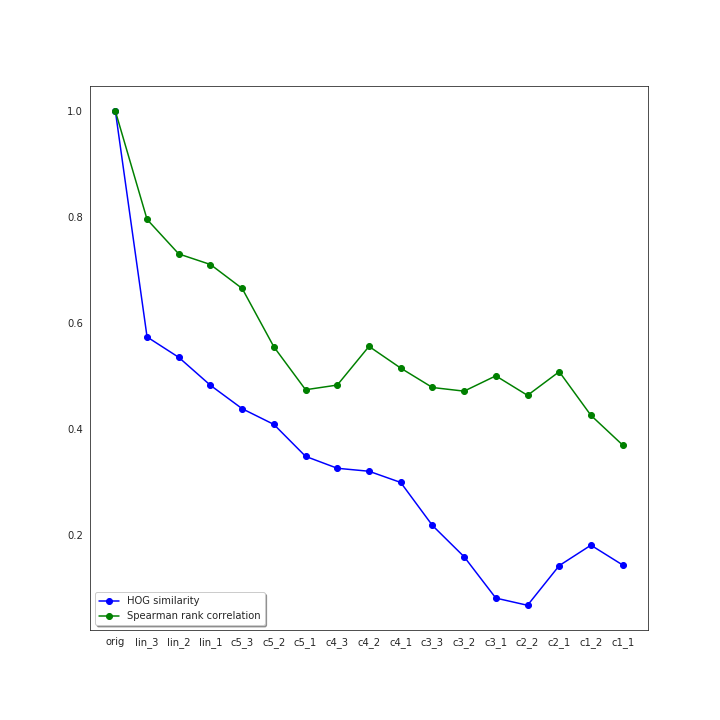}
\caption{Plot shows HOGs similarity and Spearman rank correlation between original saliency map and the maps generated by cascading-randomized VGG network. The growing difference between subsequent maps and the original one shows the sensitivity of our method to increasing level of randomization of network weights. It is a positive result confirming the sense of our method.}
\label{fig:met_casc}
\end{figure}

They proposed two type of this test: cascading randomization test and independent randomization test. In the first test type, we randomize the model weights gradually from the top layer (fully-connected layers in the CNN case) to the bottom layer (the first convolutional layer). As it was said, this process declines the learned weights in cascading manner -- from the top layers to the bottom ones. Figure \ref{fig:cascading} visualizes the cascading randomization for a few saliency methods tested in the paper \cite{sanity}. I present subsequent saliency maps obtained by LICNN for Junco bird image (Fig. \ref{fig:my_casc}) and also Pearson correlation of the histogram of gradients (HOGs) similarity and Spearman rank correlation metrics derived from comparison of original maps with subsequent maps obtained by randomize consecutive layers (Fig. \ref{fig:met_casc}). 

Here I present the comparison with the results of other methods for cascading randomization test (independent randomization results can be found in Appendix). From the plot (Fig. \ref{fig:met_casc}) we can draw a positive conclusion, because as the weights of the network are more and more random, subsequent saliency maps are more and more different from the original map -- this means that our method passes sanity check because it is sensitive to random network weights. It seems that saliency map creation method processed by LICNN is comparable with Gradient, Gradient-Input, Integrated Gradients and Grad-CAM methods.

\section{Experiments}
\subsection{Blurring images}
I decided to use the saliency maps created by LICNN to examine how much the background pixels affect the object's classification in the image. For experiments I used two datasets of images from ImageNet, each consisting of 7500 images:
\begin{itemize}
\item subset of natural images made by uniform distribution from the validation set of ILSVRC2012 competition (ImageNet images) \cite{imagenet} -- hereinafter referred to as \textit{natural images};
\item also a collection of natural images from ImageNet, but arbitrarily selected to bring neural model accuracy to decline \cite{adversarial} -- hereinafter referred to as \textit{adversarial images}.
\end{itemize}

\begin{figure}[h!]
\centering
\includegraphics[scale=0.32]{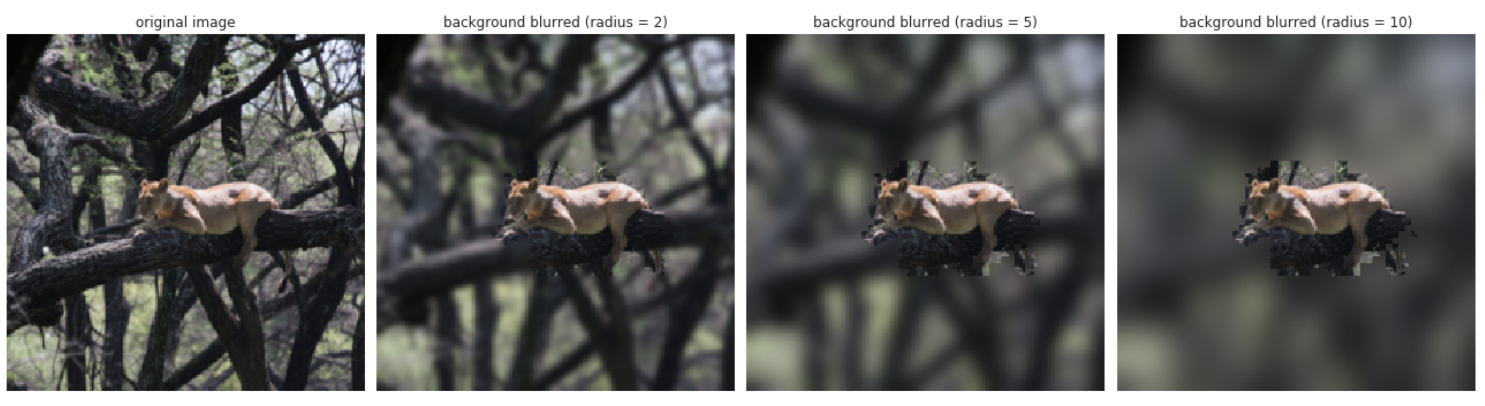}
\caption{Image with background blurred by Gaussian blurring using radius with 2, 5 and 10 values respectively.}
\label{fig:blurredbg}
\end{figure}

To neutralize the background effect in some specific way, We have obtained saliency maps for each test image\footnote{I remind that saliency maps is created by LICNN by combining top5 category-specific attention maps. The network does not receive any information about the searched class.}, then blurred the background. i.e. the regions outside of the saliency map. I used the library function for blurring from \textit{Pillow.ImageFilter} Python module which implements Gaussian blurring. I used three different blurring radius values (which determines the size of the blurring filter), which is to cause more and more image distortion, as in the example (Fig. \ref{fig:blurredbg}).

\begin{figure}[h!]
\centering
\includegraphics[scale=0.32]{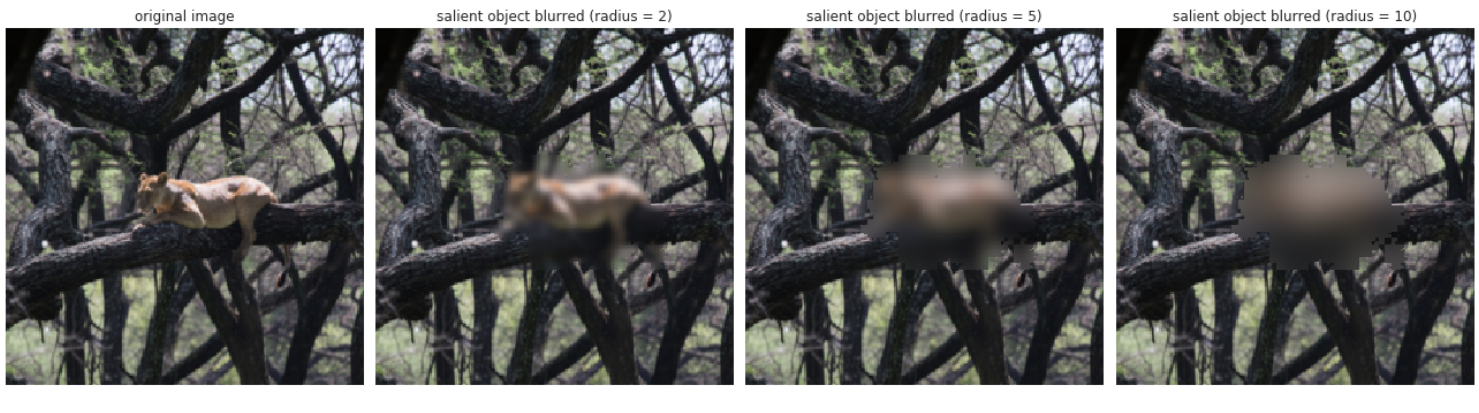}
\caption{Image with salient object blurred by Gaussian blurring using radius with 2, 5 and 10 values respectively.}
\label{fig:blurred_object}
\end{figure}

What is more, to make sure that the network responds to the foreground object, I also performed the opposite operation, i.e. blurring the saliency map area, leaving the image background unchanged (Fig. \ref{fig:blurred_object}).

\begin{figure}
    \centering
    \subfloat[natural images]{{\includegraphics[width=6cm]{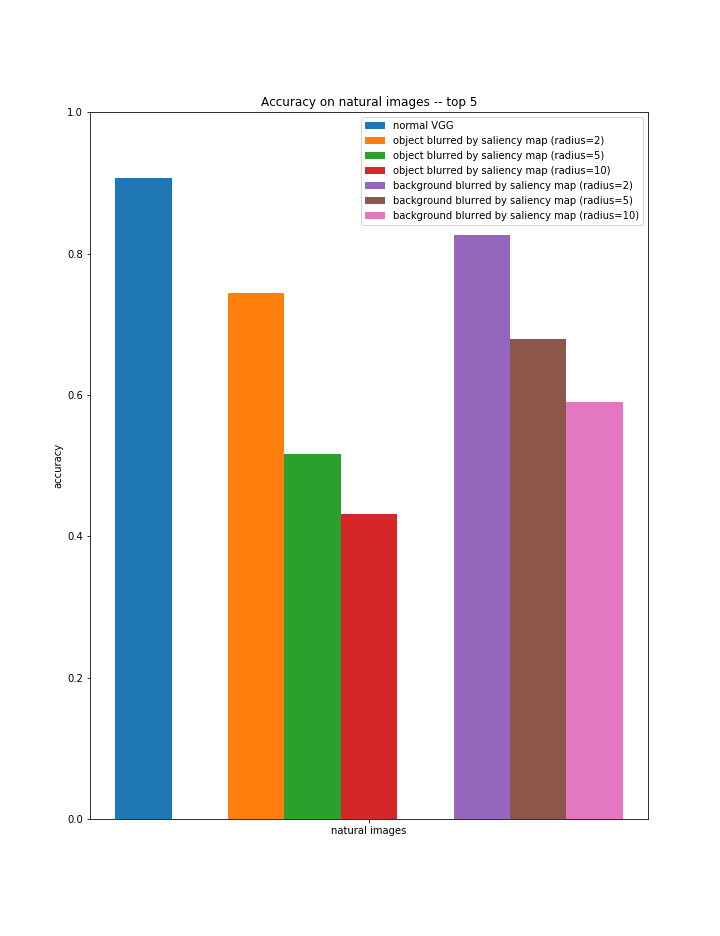} }}
    \qquad
    \subfloat[adversarial images]{{\includegraphics[width=6cm]{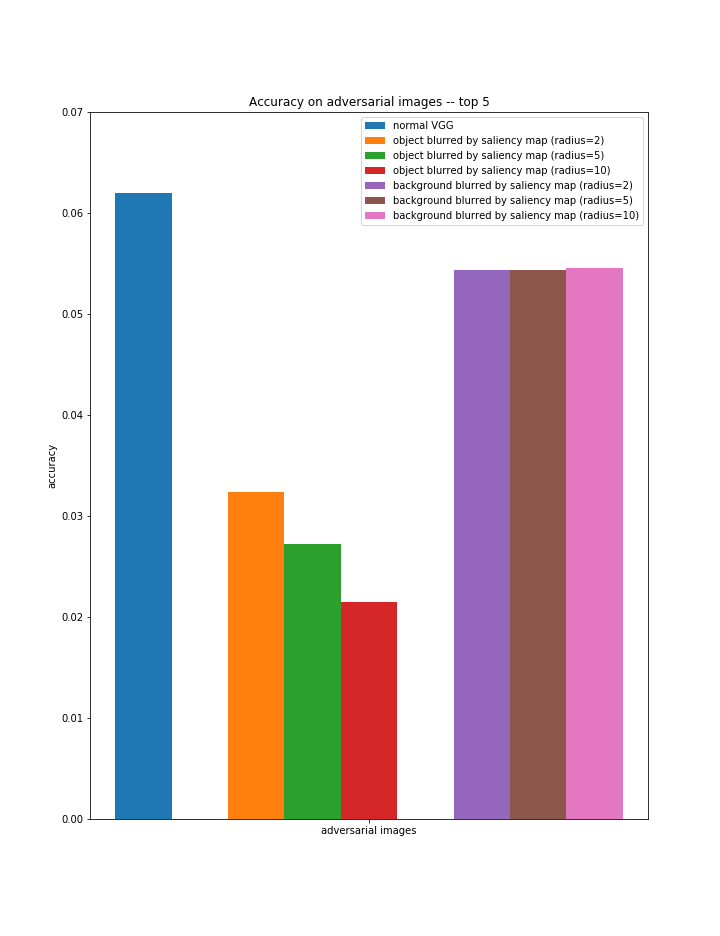} }}
    \caption{Top5 accuracy on natural and adversarial images.}
    \label{fig:top5}
\end{figure}

Using the VGG-16 network for classification, I calculated the top5 accuracy of this network on natural and adversarial images datasets with the following variants: original, with blurred salient object and with blurred background (with increasing blur) (Fig. \ref{fig:top5}).
As can be seen on the diagram (Fig. \ref{fig:top5}(a)), on the natural images dataset, along with the background blur, the network accuracy decreases. If we assume that saliency maps precisely cover important objects in the image, then the aforementioned decrease in accuracy indicates the influence of background pixels on the classifier predictions. In addition, it is worth noting that in the case of significant object blurring, the network accuracy is NOT drastically reduced and with complete blurring of the main object it still amounts as much as half of the network accuracy on unmodified images. This behavior also seems to support the hypothesis that background pixels can have a significant impact on classifier execution.

On the diagram showing network accuracy on adversarial images (Fig. \ref{fig:top5}(b)) we notice that despite increasing background blur, accuracy does not decrease (and even increases slightly for radius = 10). This prompted me to look more closely and analyze the adversarial images which were correctly classified by the network when they had a blurred background, but incorrectly for the original (unmodified) state. I describe the observations in the next section.

\begin{figure}[h!]
\centering
\includegraphics[scale=0.32]{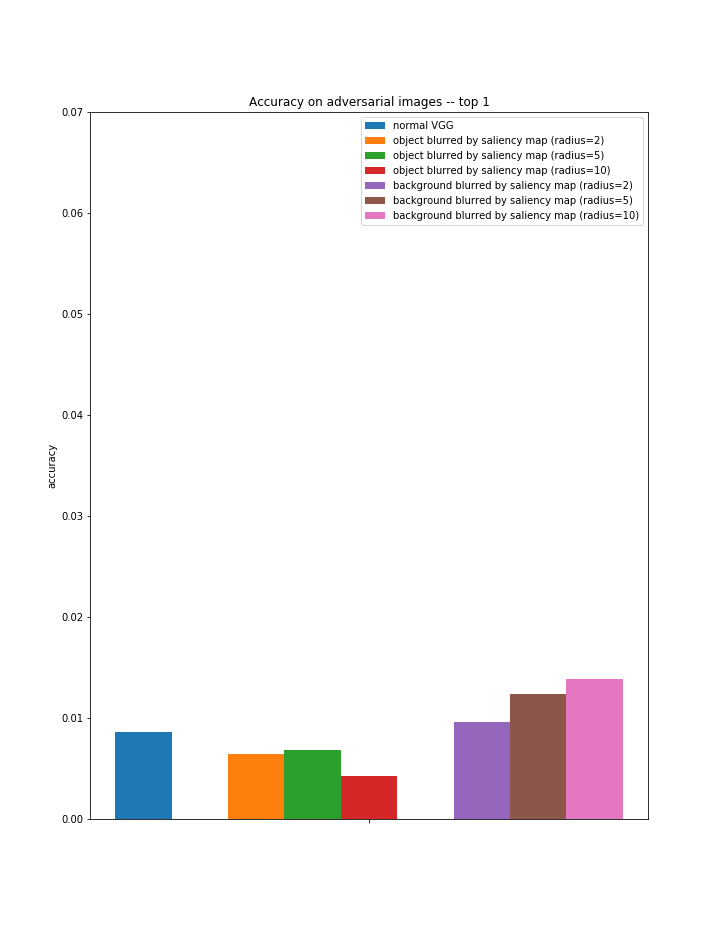}
\caption{Top1 accuracy on adversarial images.}
\label{fig:top1}
\end{figure}

We could observe a surprising increase in accuracy caused by background blurring on diagram showing top1 accuracy on adversarial images (Fig. \ref{fig:top1}). Such increase ideally shows the disastrous background effect on network predictions on adversarial images, but since the change in accuracy is very small and this trend is not maintained in top5, it is difficult to consider this case sufficient to confirm the overall trend.

\subsection{Types of background influences}

\begin{figure}[h!]
\centering
\includegraphics[scale=0.27]{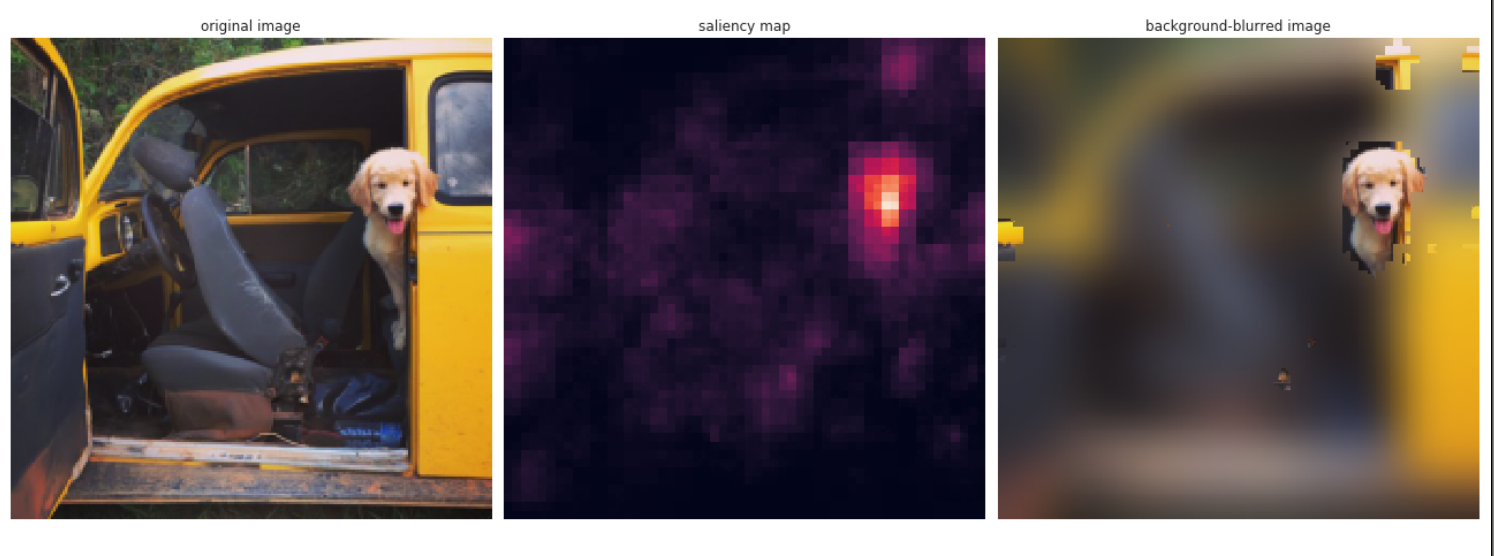}
\caption{Example of background impact on image classification. Normally classified as: (1) taxi, (2) school bus, (3) minivan, (4) minibus, (5) pickup; whereas blurred image as: (1) GOLDEN RETRIEVER, (2) school bus, (3) Appenzeller (dog breed), (4) EntleBucher (dog breed), (5) Cardigan Welsh corgi (also dog breed). Ground truth is indicated by BIG letters.}
\label{fig:taxidog}
\end{figure}

After analyzing adversarial images that were normally misclassified by the network and classified correctly, when they had a blurred background\footnote{Mostly I looked at images blurred using radius=10.}, I identified some types of background effects on the object in the image that may confuse the network as to the object's true category. I present them below:

1. The background can be an independent category which, because it occupies a significantly larger area of the image, has such a great impact on the network that the signal about the object occupying less space is not noticed. After looking at the generated saliency map (Fig. \ref{fig:taxidog}), it seems that the lateral inhibition method primarily leaves close clusters of corresponding neurons with a strong signal, which can cause weak and diffused signals from the background to be weakened or suppressed (strongly marked dog head on the map, poorly taxi background). Figure \ref{fig:taxidog} is normally classified as\footnote{Order from the highest score.}: (1) taxi, (2) school bus, (3) minivan, (4) minibus, (5) pickup; whereas blurred image as: (1) GOLDEN RETRIEVER, (2) school bus, (3) Appenzeller (dog breed), (4) EntleBucher (dog breed), (5) Cardigan Welsh corgi (also dog breed).

\begin{figure}[h!]
\centering
\includegraphics[scale=0.27]{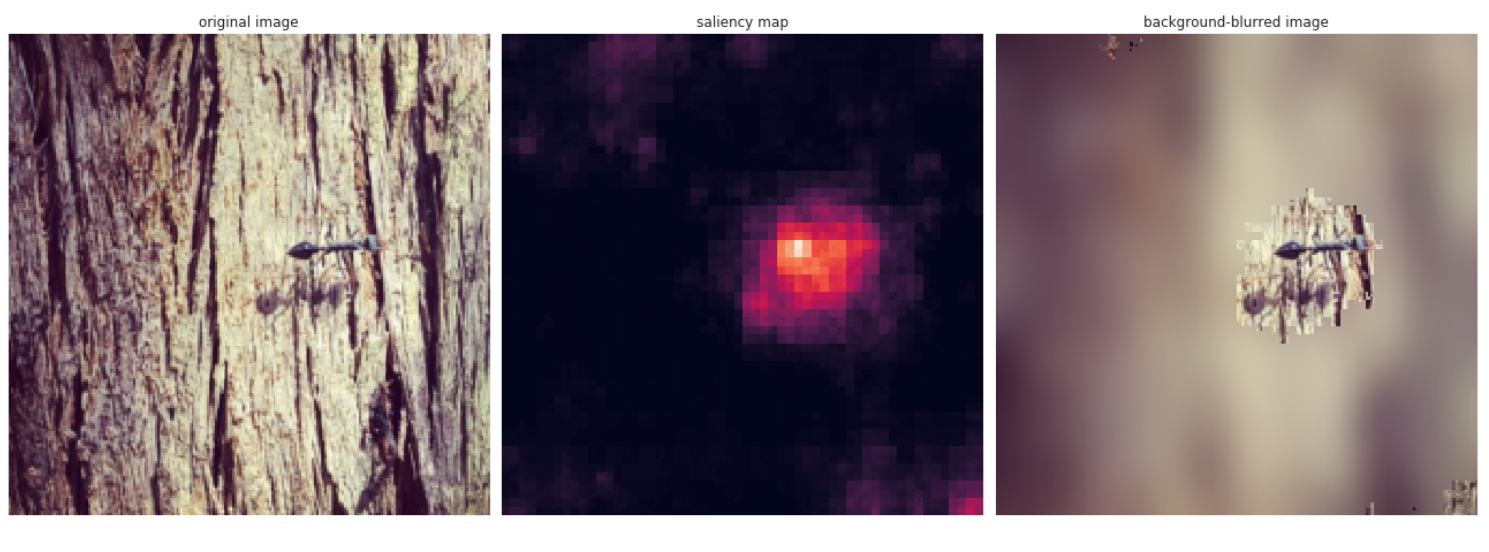}
\caption{Example of background impact on image classification. Normally classified as:  (1) nail, (2) agama, (3) walking stick, (4) long-horned beetle, (5) foxsquirrel;  whereas  blurred  image  as:  (1)  barn  spider,  (2)  ANT,  (3)  tick  (4)  spaceshuttle, (5) cardoon. Ground truth is indicated by BIG letters.}
\label{fig:nailant}
\end{figure}

2. In this case, the background contextually suggests a different object than the right one, because either the searched-class object has an unnatural background, or another class occurs much more often against this background. Figure \ref{fig:nailant} normally classified as: (1) nail, (2) agama, (3) walking stick, (4) long-horned beetle, (5) fox squirrel; whereas blurred image as: (1) barn spider, (2) ANT, (3) tick (4) space shuttle, (5) cardoon.

For each type of background influences, I counted dozens of cases from the analyzed images. A few more examples can be found in Appendix.

\begin{figure}[h!]
\centering
\includegraphics[scale=0.27]{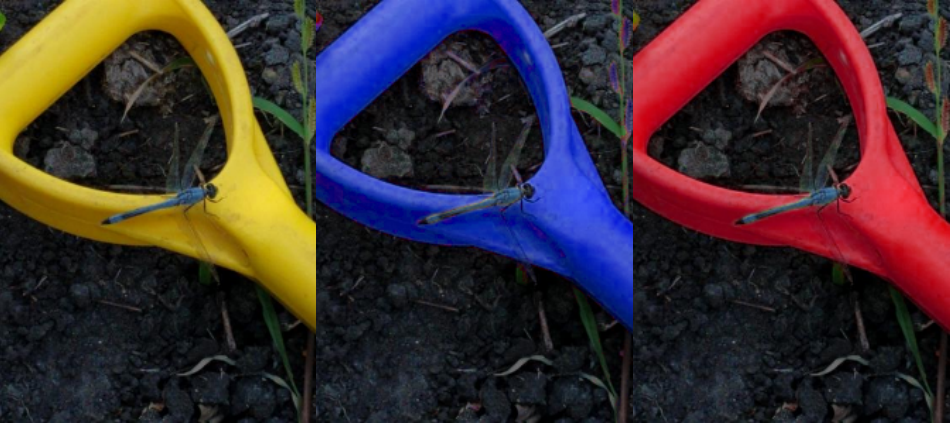}
\caption{Interesting example from the adversarial images paper\cite{adversarial}. When the shovel is yellow, image is misclassified as "banana" and in the other cases classified correctly as "dragonfly".}
\label{fig:dragonfly}
\end{figure}

\section{Conclusions}
In this work, I have implemented LICNN model which adopts lateral inhibition -- neurobiological concept of mutual neuronal suppression -- for creating high-quality saliency maps. As we could see they emphasize relevant objects well highlighting discriminative and salient features but they also have some shortcomings  as in Fig. \ref{fig:no_people_img}. Nevertheless, they are a helpful tool for expressing input-output relationships in a neural network  and provide new opportunities to improve them. In this work, saliency maps were used to examine the background impact on the predictions of a classification network.

The conducted experiments show that background pixels have a visible impact on results in a neural network. Probably this work could be further developed in the direction of training a discriminative network (with saliency maps help), which would not be so much affected by the background. I hope that this work will contribute to increasing awareness in the scientific world of the impact of background on network predictions and will cause a thorough examination of this topic to improve network performance.

\newpage


\newpage

\section{Appendix}

\textbf{The kind of attention map I used} \\
In this work I used response-based attention map kind from the LICNN paper, because I suggested by authours' statement that "the gradient-based attention maps are essentially generated by measuring the contribution of activated patterns, which account for selective attention, while the response-based attention maps depend on the strength of input patterns, which are more related to saliency" \cite{cao}. As the authors of the paper I involved response-based attention map to a saliency task.

\textbf{Independent randomization test results} \\

\begin{figure}[h!]
\centering
\includegraphics[scale=0.19]{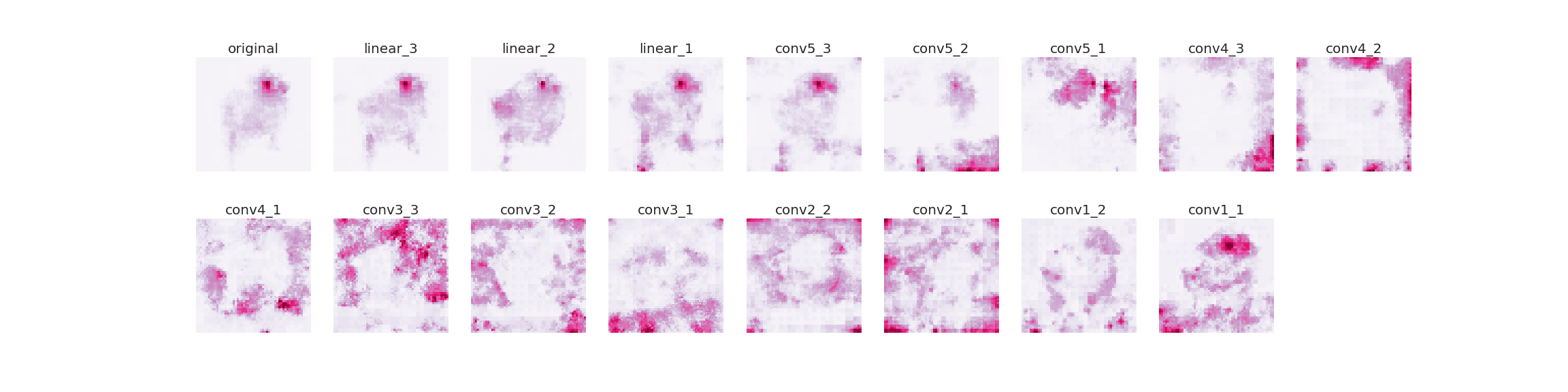}
\caption{Independent randomization resulting saliency maps of the Junco bird image.}
\label{fig:indep_rand}
\end{figure}
\begin{figure}[h!]
\centering
\includegraphics[scale=0.4]{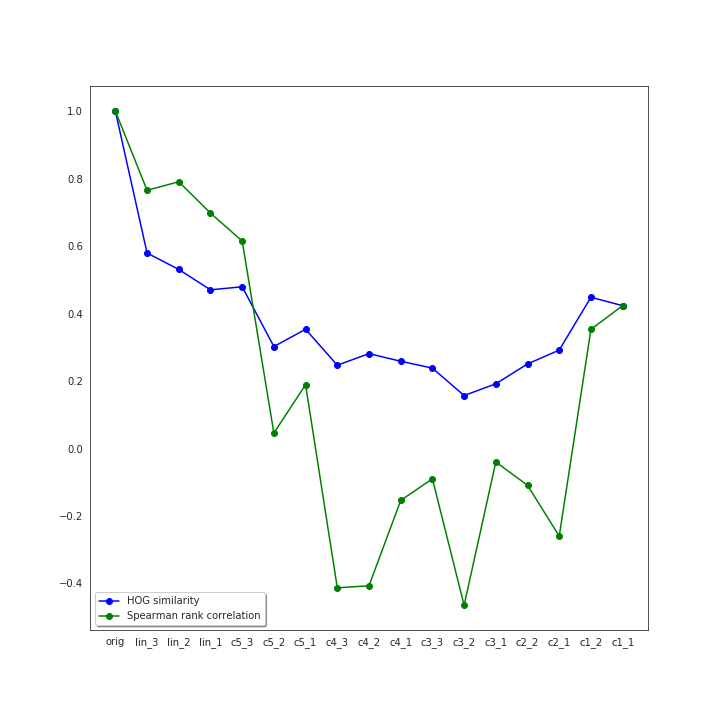}
\caption{Plot shows HOGs similarity and Spearman rank correlation between original saliency map and the maps generated by independent-randomized VGG network.}
\label{fig:met_indep}
\end{figure}

\textbf{Back-propagation after softmax} \\
I perform back-propagation after softmax instead of before it (here \cite{bib:saliency} they do the opposite, but I do not understand why). My experiments show that when doing back-propagation after softmax, category-specific attention map is more focused on its category. In the LICNN paper \cite{cao} there is no information regarding the place when we should perform back-propagation (before or after softmax). In the following figures I present top5 category-specific attention maps for two images created by my LICNN implementation, they differ in location of back-propagation execution. It seems that additional research is needed to clarify the relationship between the place of back-propagation performing and the resulting maps.

\begin{figure}[h!]
\centering
\includegraphics[scale=0.36]{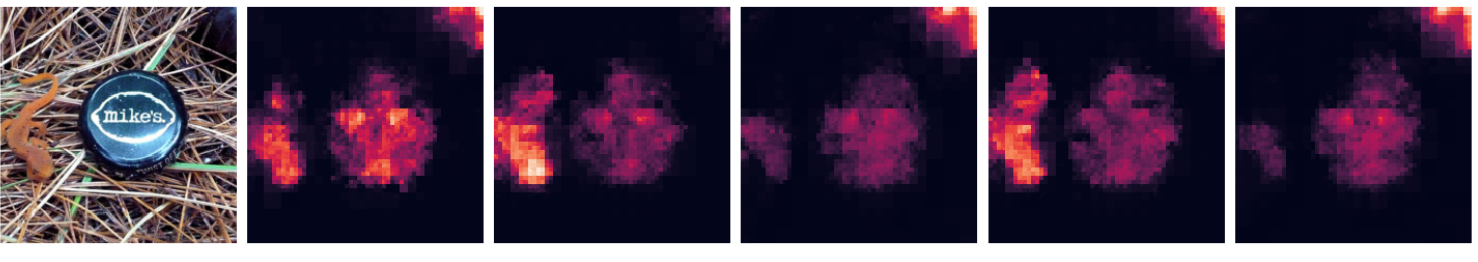}
\caption{Back-propagation after softmax. Top5 predicted classes are: (1) bottlecap, (2) eft, (3) dung beetle, (4) ringneck snake, (5) ground beetle.}
\label{fig:soft_aft_1}
\end{figure}
\begin{figure}[h!]
\centering
\includegraphics[scale=0.36]{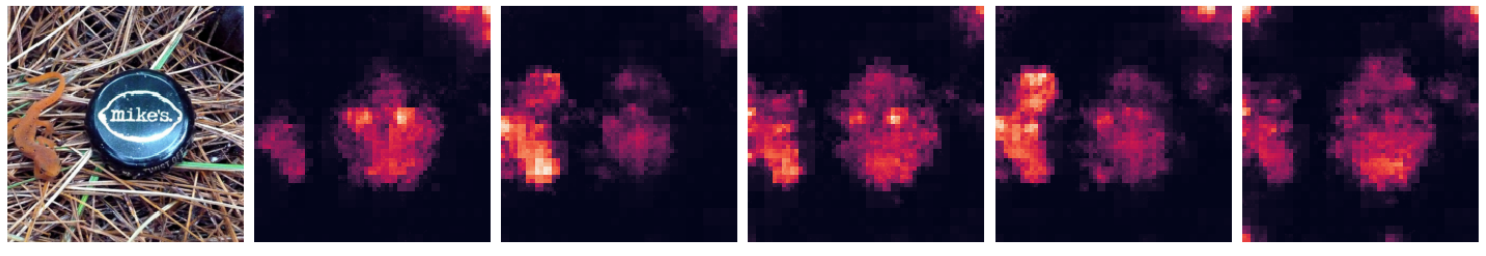}
\caption{Back-propagation before softmax. Top5 predicted classes are: (1) bottlecap, (2) eft, (3) dung beetle, (4) ringneck snake, (5) ground beetle.}
\label{fig:soft_bef_1}
\end{figure}

\begin{figure}[h!]
\centering
\includegraphics[scale=0.36]{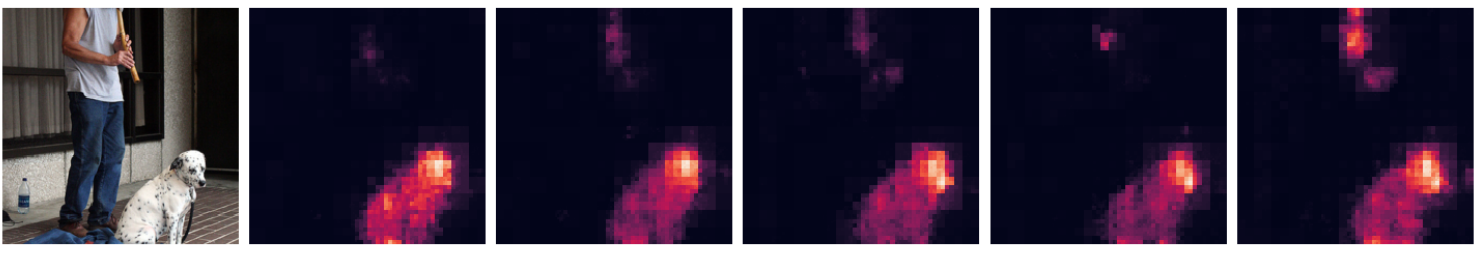}
\caption{Back-propagation after softmax. Top5 predicted classes are: (1) dalmatian, (2) English setter, (3) Great Dane, (4) whippet, (5) German short-haired pointer.}
\label{fig:soft_aft_2}
\end{figure}

\begin{figure}[h!]
\centering
\includegraphics[scale=0.36]{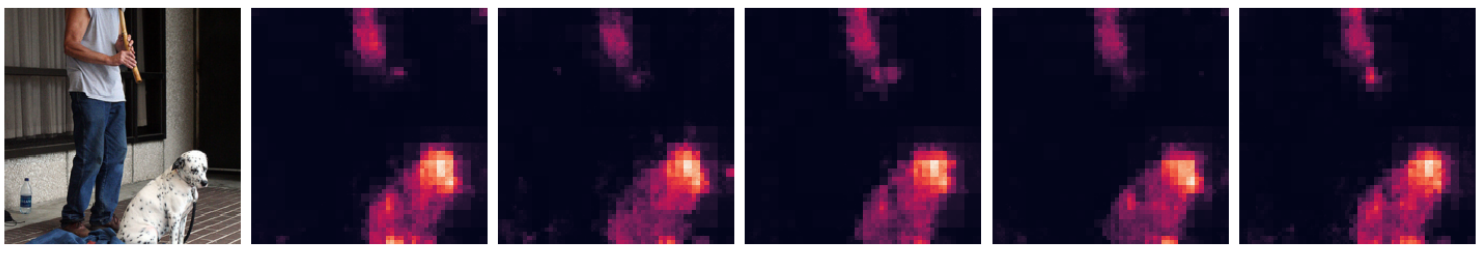}
\caption{Back-propagation before softmax.  Top5 predicted classes are: (1) dalmatian, (2) English setter, (3) Great Dane, (4) whippet, (5) German short-haired pointer.}
\end{figure}

\newpage

\textbf{More background impact examples} \\
\begin{figure}[h!]
\centering
\includegraphics[scale=0.36]{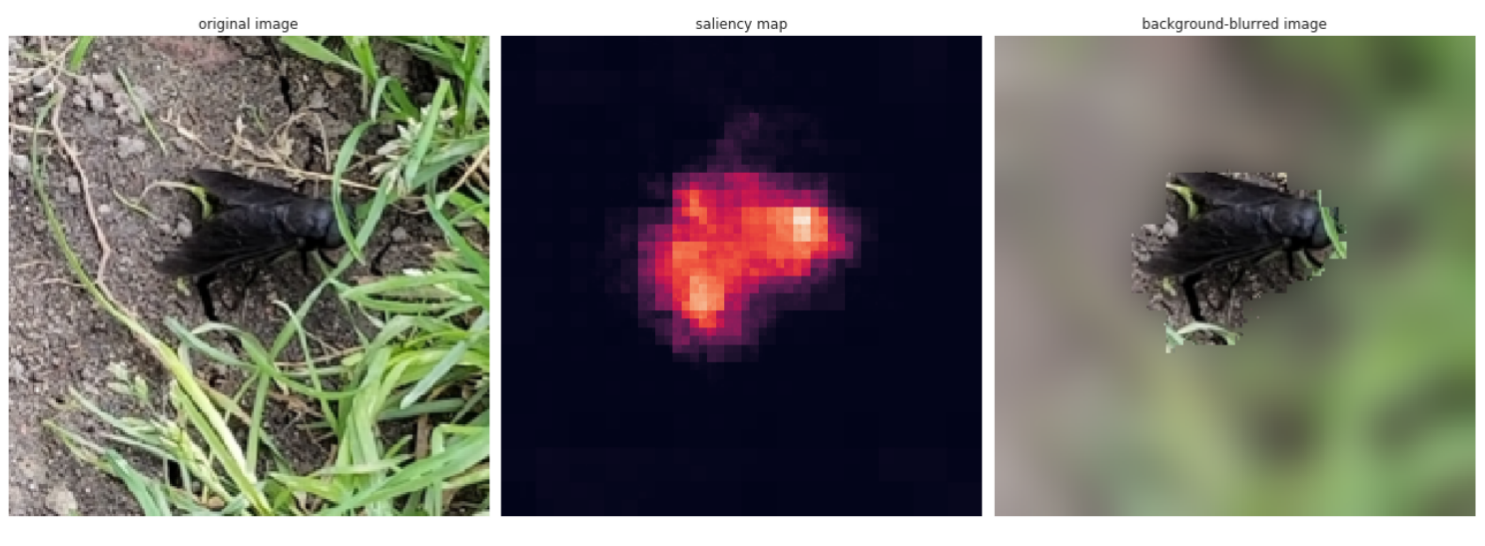}
\caption{Top5 classes predicted on unmodified image are: (1) dung beetle, (2) ground beetle, (3) mud turtle, (4) terrapin, (5) rhinoceros beetle. Top5 classes on blurred image: (1) FLY, (2) ground beetle, (3) cicada, (4) leafhopper, (5) leaf beetle. Ground truth is indicated by BIG letters.}
\end{figure}

\begin{figure}[h!]
\centering
\includegraphics[scale=0.36]{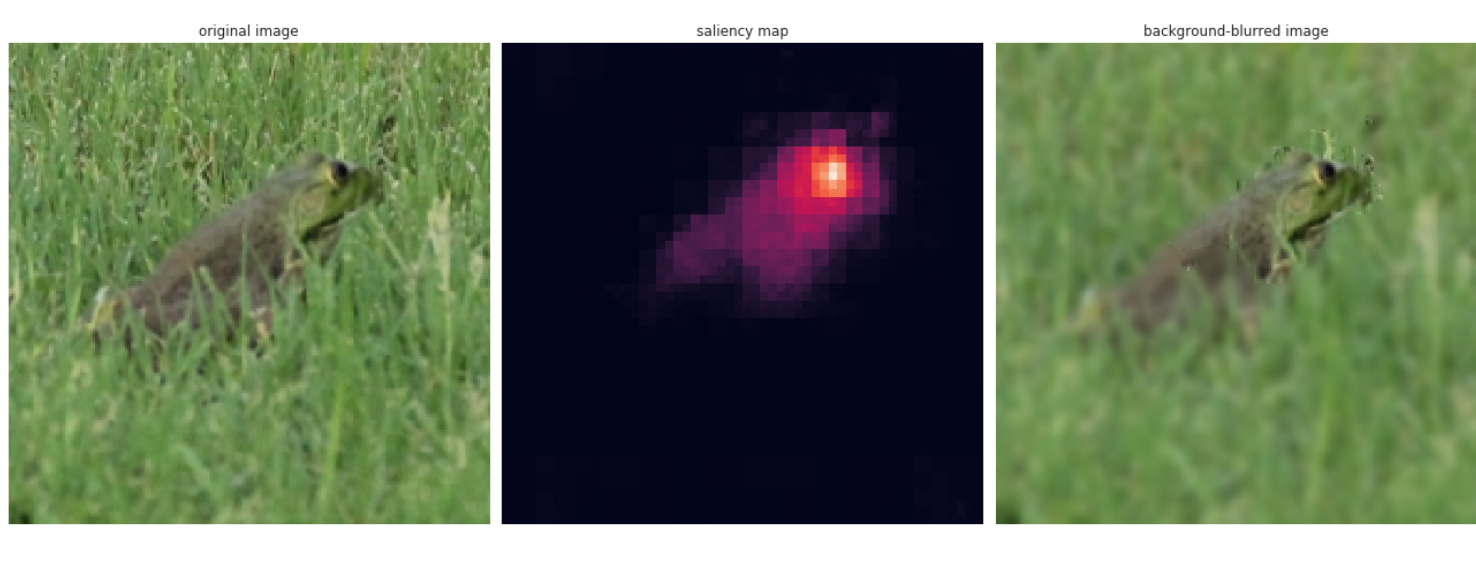}
\caption{Top5 classes predicted on unmodified image are: (1) fox squirrel, (2) wood rabbit, (3) hare, (4) mongoose, (5) marmot. Top5 classes on blurred image: (1) tailed frog, (2) fox squirrel, (3) African chameleon, (4) wood rabbit, (5) BULLFROG. Ground truth is indicated by BIG letters.}
\end{figure}

\begin{figure}[h!]
\centering
\includegraphics[scale=0.36]{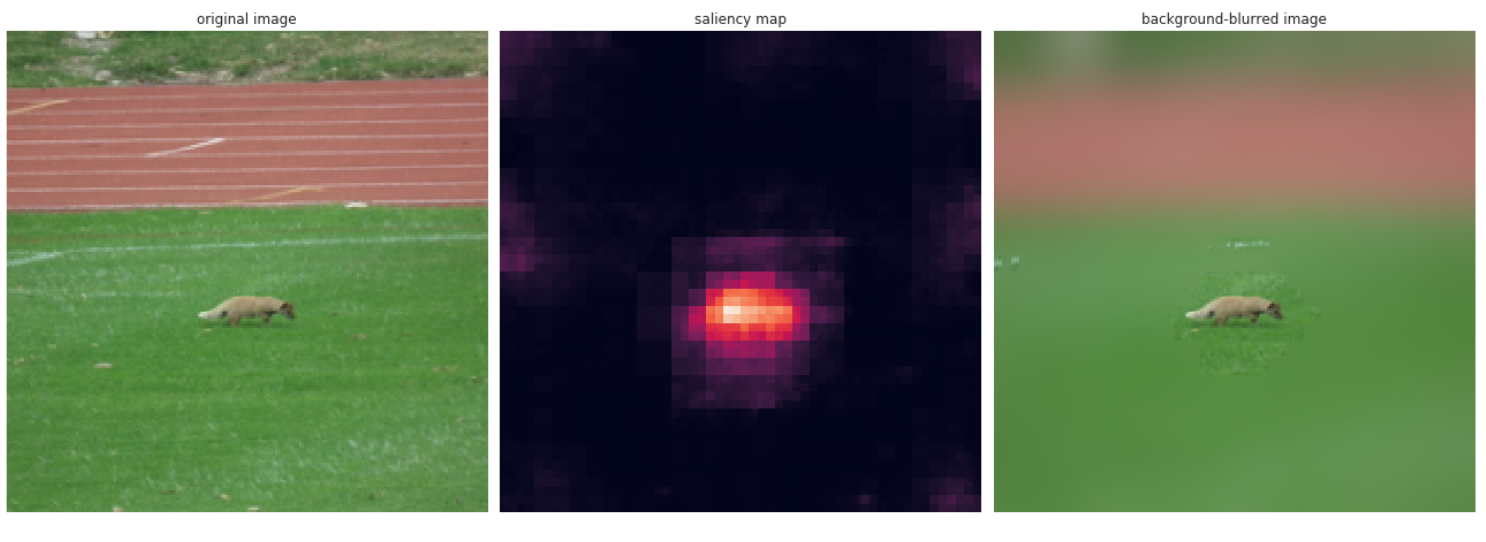}
\caption{Top5 classes predicted on unmodified image are: (1) ballplayer, (2) malinois, (3) Rhodesian ridgeback, (4) brush kangaroo, (5) Hungarian pointer. Top5 classes on blurred image: (1) platypus, (2) fox squirrel, (3) MONGOOSE, (4) hare, (5) dugong. Ground truth is indicated by BIG letters.}
\end{figure}

\end{document}